\documentclass[11pt]{article}
\usepackage{acl}
\usepackage{times}
\usepackage{latexsym}
\usepackage{amsmath}
\usepackage{amssymb}
\usepackage{booktabs}
\usepackage{multirow}
\usepackage{graphicx}
\usepackage{xcolor}
\usepackage{subcaption}
\usepackage{float}

\newcommand{\neff}{n_{\text{eff}}}
\newcommand{\phiavg}{\bar{\phi}}

\title{Nine Judges, Two Effective Votes: \\ Correlated Errors Undermine LLM Evaluation Panels}

\author{
  Guneet Kohli \\
  Apple \\
  \texttt{g\_kohli@apple.com}
}

\date{}

\begin{document}
\maketitle

\begin{abstract}
LLM-as-a-judge panels aggregate votes from multiple models, with the expectation that diverse models yield more reliable evaluations. We develop a framework to measure the true informational value of such panels and quantify how far their reliability falls short of the independent-voting ideal.
Testing a panel of 9 frontier LLMs from 7 model families on three natural language inference datasets (each with 100 human annotations per item), we find that the 9 judges effectively provide only about 2 independent votes' worth of information. Roughly three-quarters of the panel's nominal independence is lost because the models make the same mistakes on the same items.
The consequences are stark: the panel's actual accuracy falls 8--22 percentage points short of what independent voting would achieve, and the best single judge matches or outperforms the full panel across all conditions. Neither adding more judges nor using smarter aggregation algorithms helps --- established methods close at most 11\% of this gap, even with access to the correct answers.
We quantify these findings using the Kish effective sample size ($\neff$) and a Condorcet null model, and show the deficit is robust across prompt variants, temperatures, chain-of-thought reasoning, and a pairwise preference task (RewardBench). The bottleneck is correlated judges, not the aggregation algorithm, implying that scaling up panels cannot substitute for genuinely independent evaluation.
\end{abstract}

\section{Introduction}

LLM-as-a-judge evaluation has become a standard methodology for scalable assessment of language model outputs \citep{zheng2023judging}. To mitigate single-model biases, researchers have turned to multi-model panels --- ensembles of diverse LLMs that vote on evaluation items --- with the expectation that cross-model diversity yields something approaching independent assessment \citep{verga2024replacing}. The intuition is compelling: if models from different providers, trained on different data, make different errors, then majority voting should be robust.

This intuition draws on the logic of the Condorcet Jury Theorem \citep{condorcet1785}: if each voter is better than chance and votes independently, majority-vote accuracy increases monotonically with panel size and approaches certainty. The practical appeal is clear --- adding more judges should always help, and a panel of 9 should be far more reliable than any single judge.

But is it? We evaluate a 9-judge panel spanning 7 model families on three natural language inference (NLI) benchmarks --- MNLI, SNLI, and AlphaNLI --- each with 100 human annotations per item (\S3). The panel provides negligible or negative lift over its single best member. On MNLI, the panel (72.0\%) barely edges the best judge (Qwen3-32B, 71.8\%) by 0.2pp --- within noise; on SNLI, the best judge dominates (Claude Sonnet 4.5, 84.2\% vs.\ panel 77.7\%); and on AlphaNLI, an abductive reasoning task with a different label set, the pattern persists (91.2\% vs.\ 88.7\%). These results are impossible under the independence assumption but expected when errors are highly correlated. Recent work has documented such correlated errors across LLMs on standard benchmarks \citep{kim2025correlated}, and conceptual arguments suggest that shared training paradigms should induce dependence \citep{lefort2024examining}. However, no prior work has \emph{quantified} the effective independence of LLM judge panels in a way that directly connects to majority-vote reliability, using a ground truth rich enough to validate the measurement.

We address this gap with three contributions:

\begin{enumerate}
    \item \textbf{A diagnostic framework for LLM judge panels.} We combine the Kish effective sample size ($\neff$) --- a measure of how many truly independent votes a panel contains --- with a Condorcet null model that simulates what majority-vote accuracy \emph{would be} if judges voted independently. Applied to a 9-judge, 7-family panel on three ChaosNLI datasets \citep{nie2020chaosnli}, we find $\neff \approx 2.0$--$2.5$: the panel contains roughly two independent votes worth of information. The accuracy shortfall relative to this independent prediction (the \emph{Condorcet gap}) is 8--22 percentage points (pp; permutation $p < 10^{-4}$).

    \item \textbf{A severe independence deficit, stable across tasks, prompts, and temperatures.} The deficit is remarkably consistent: $\neff \approx 2.0$--$2.5$ across three NLI datasets, three prompt variants, two temperature settings, and a pairwise preference task (RewardBench; $\neff = 2.0$) --- despite panel accuracy ranging from 69\% to 93\%. Across all conditions, the panel fails to meaningfully outperform the single best judge. The scaling curve shows that adding judges beyond 5 yields negligible benefit, with effective independence asymptoting at roughly 2.3--3.1 (varying by dataset).

    \item \textbf{A negative result: aggregation cannot overcome correlation.} Established aggregation methods --- Dawid-Skene EM \citep{dawid1979maximum} and accuracy-weighted voting --- close at most 11\% of the Condorcet gap across all four datasets, even with oracle access to gold labels. The bottleneck is correlated inputs, not the algorithm: no weighting scheme can extract a third independent vote from $\sim$2.2 effective votes of information.
\end{enumerate}

These results have direct practical implications: paying for 9 opinions but receiving the informational equivalent of $\sim$2 is a substantial inefficiency. The marginal value of additional judges is near zero, and unanimous panel agreement is far less diagnostic than it appears. The path forward is not larger ensembles of similar models, but diversification of the underlying reasoning --- models that genuinely differ in how they process information.

\section{Related Work}

\paragraph{LLM-as-a-judge.}
\citet{zheng2023judging} established the LLM-as-a-judge paradigm, and subsequent work has revealed systematic biases \citep{wang2024large, ye2024justice, thakur2024judging} and raised measurement-theory concerns about validity and reliability \citep{chehbouni2025neither, calderon2025alternative}. Our work goes beyond cataloguing individual biases to quantify the \emph{structural} dependence among judges --- a more fundamental constraint on panel reliability.

\paragraph{LLM judge panels.}
\citet{verga2024replacing} proposed PoLL (Panel of LLM Evaluators), demonstrating that panels of smaller, diverse models outperform single large judges across six datasets. Importantly, PoLL compares panels to the \emph{average} individual judge, where panels naturally win by diversifying away individual quirks. Our finding --- that the \emph{best} individual outperforms the panel --- does not contradict PoLL but reveals a different phenomenon: when judges are highly correlated, majority voting dilutes the best judge's signal with redundant weaker votes. Our work complements PoLL by showing that the panel's effective information content is far lower than the raw panel size suggests.

The Trust-or-Escalate framework \citep{jung2025trust} provides provable guarantees using \emph{single-model} confidence to decide when to escalate to human review. Our approach differs fundamentally: we use \emph{cross-model} disagreement and show that this disagreement is itself unreliable due to correlated errors.

\paragraph{Correlated errors and scaling limits.}
Most closely related to our work, \citet{kim2025correlated} conducted a large-scale study of error correlation across 350+ LLMs, finding that models agree on wrong answers 60\% of the time on some benchmarks. \citet{jiang2025hivemind} demonstrated what they term the ``artificial hivemind'' effect: LLMs produce strikingly homogeneous outputs on open-ended tasks, both within and across model families, and LLM judges are poorly calibrated on items where human annotators disagree --- our $\neff \approx 2.0$--$2.4$ provides a precise quantification of this qualitative insight. \citet{dorner2025limits} proved a complementary theoretical result: when the judge is no more capable than the evaluated model, no debiasing method can reduce the required ground-truth data by more than half, establishing a fundamental scaling ceiling. We build on this body of work by (a) measuring effective independence in the LLM-as-a-judge setting using Kish $\neff$, (b) quantifying the Condorcet gap that correlation creates, and (c) showing that established aggregation methods cannot close this gap.

\paragraph{Condorcet Jury Theorem and ensembles.}
The Condorcet Jury Theorem \citep{condorcet1785} underpins much of the intuition behind ensemble methods \citep{dietterich2000ensemble, surowiecki2004wisdom}: diverse, independent voters collectively outperform individuals. \citet{austen1996information} showed that independence is necessary, not just sufficient --- correlated voters can perform \emph{worse} than individuals. In the LLM setting, \citet{lefort2024examining} applied the Condorcet Jury Theorem to sentiment analysis ensembles, finding marginal improvement consistent with a lack of independence. \citet{turkmen2026information} formalized this via an information-theoretic error floor. Our work empirically validates these theoretical concerns, providing the first item-level measurement of effective independence in a judge evaluation setting.

\paragraph{Statistically principled aggregation.}
\citet{zhao2025care} proposed CARE, a confounder-aware aggregation framework that models inter-judge correlations, reducing aggregation error by up to 25\%. The crowdsourcing literature offers related methods: the Dawid-Skene model \citep{dawid1979maximum} estimates annotator error rates via EM, and \citet{raykar2010learning} extended this to learning from noisy crowds. Where these methods propose \emph{solutions} (better aggregation), our paper provides both the \emph{diagnosis} --- quantifying how much independence is actually present --- and a \emph{negative result}: even with oracle access to gold labels, established aggregation methods close at most 11\% of the Condorcet gap (\S\ref{sec:aggregation}), suggesting that the problem is structural rather than algorithmic.

\paragraph{Human label variation and ChaosNLI.}
Human disagreement on NLI items is systematic, not mere noise \citep{pavlick2019inherent, plank2022problem}. \citet{nie2020chaosnli} created ChaosNLI with 100 annotator labels per item, built on MNLI \citep{williams2018broad}, providing the richest available ground truth for studying disagreement patterns. \citet{lee2023dissenting} showed that single LLMs fail to capture the distributional properties of human disagreement on ChaosNLI. We extend this line by testing whether \emph{multi-model panels} can capture human disagreement patterns, finding that correlated errors severely limit their ability to do so.

\section{Methodology}

\subsection{Datasets}

We use ChaosNLI \citep{nie2020chaosnli}, which provides 100 annotator labels per item. Our primary dataset is ChaosNLI-MNLI (1{,}599 MNLI items; \citealt{williams2018broad}), with labels entailment~(e), neutral~(n), or contradiction~(c). We replicate on ChaosNLI-SNLI (1{,}514 SNLI items; \citealt{bowman2015large}) as a same-task robustness check, and on ChaosNLI-AlphaNLI (1{,}532 abductive NLI items; \citealt{bhagavatula2020abductive}) as a cross-task replication with a different label set (2-class: hypothesis~1 vs.\ hypothesis~2) and reasoning type (abductive rather than textual entailment). From each dataset, we sample 1{,}000 items using entropy-stratified sampling (equal proportions from low, medium, and high human-entropy terciles) with seed 42.

The gold standard for each item is the majority vote of 100 annotators (tie-breaking details in Appendix~\ref{app:tiebreaks}). Human entropy (Shannon entropy, base-2) ranges from 0.00 to 1.58 bits on MNLI/SNLI and 0.00 to 1.00 bits on AlphaNLI (lower maximum due to 2-class), providing rich ground truth against which to validate panel behavior. We present MNLI results in the main body and report SNLI and AlphaNLI replication results in \S\ref{sec:snli}.

\subsection{Judge Panel}

Our panel consists of 9 judges from 7 model families (Table~\ref{tab:judges}). All judges use temperature 0.0 and receive a standardized NLI classification prompt (Appendix~\ref{app:prompt}). Rare parse failures ($<$0.1\% of all judgments; 21 of 28 from Llama 4 Maverick, with 5 from Gemini 2.5 Pro and 2 from Claude Sonnet 4.5) are handled via deterministic hash-based random assignment to \{e, n, c\} to avoid systematic bias. With an odd number of judges, majority-vote ties are eliminated on 2-class tasks (AlphaNLI, RewardBench); on 3-class tasks (MNLI, SNLI), the rare remaining ties (0.4--1.1\% of items) are broken via deterministic SHA-256 hashing of the item index and vote sequence, ensuring reproducibility (Appendix~\ref{app:tiebreaks}).

\begin{table}[t]
\centering
\small
\begin{tabular}{lll}
\toprule
\textbf{Judge} & \textbf{Family} & \textbf{Error Rate} \\
\midrule
GPT-4o & OpenAI & 0.354 \\
GPT-4o-mini & OpenAI & 0.356 \\
Claude Sonnet 4.5 & Anthropic & 0.317 \\
Gemini 2.5 Pro & Google & 0.324 \\
Llama 4 Maverick & Meta & 0.299 \\
Llama 4 Scout & Meta & 0.332 \\
Qwen3-32B & Alibaba & 0.282 \\
Mistral Large 3 & Mistral & 0.338 \\
DeepSeek-V3 & DeepSeek & 0.321 \\
\bottomrule
\end{tabular}
\caption{Judge panel: 9 judges from 7 model families. Error rates are computed on ChaosNLI-MNLI against the 100-annotator majority label. The panel does not meaningfully outperform the best individual judge on any dataset (Table~\ref{tab:crossdataset}).}
\label{tab:judges}
\end{table}

\subsection{Effective Sample Size ($\neff$)}
\label{sec:neff}

We measure effective independence using two complementary approaches.

\paragraph{Kish design-effect $\neff$.}
For each judge, we construct a binary error vector $\mathbf{e}_j \in \{0,1\}^{1000}$ where $e_{j,i} = 1$ if judge $j$ disagrees with the ChaosNLI majority label on item $i$. We compute the pairwise phi coefficient $\phi_{jk}$ between all $\binom{9}{2} = 36$ judge pairs, then apply the Kish formula:
\begin{equation}
    \neff = \frac{k}{1 + (k-1)\phiavg}
    \label{eq:kish}
\end{equation}
where $k$ is the number of judges and $\phiavg = \frac{1}{\binom{k}{2}}\sum_{j<k}\phi_{jk}$ is the mean pairwise correlation \citep{kish1965survey}. For binary error vectors, the phi coefficient reduces to the Pearson product-moment correlation, which is the quantity the Kish formula requires \citep{kish1965survey}. Alternative association measures (e.g., Cohen's kappa) conflate prevalence with dependence; phi isolates the linear dependence that directly degrades majority-vote performance. This formula assumes exchangeability (approximately equal pairwise correlations); we validate this assumption against the eigenvalue method below.

\paragraph{Eigenvalue $\neff$.}
As a robustness check that does not assume exchangeability, we compute $\neff^{\text{eigen}} = k / \lambda_{\max}$, where $\lambda_{\max}$ is the largest eigenvalue of the $k \times k$ phi correlation matrix \citep[cf.][]{bretherton1999effective}. Under perfect independence, $\lambda_{\max} = 1$ and $\neff^{\text{eigen}} = k$; under perfect correlation, $\lambda_{\max} = k$ and $\neff^{\text{eigen}} = 1$.

\paragraph{Bootstrap confidence interval.}
We resample the 1{,}000 items with replacement 10{,}000 times, recomputing $\neff$ (Kish) for each resample, and report the 2.5th--97.5th percentile interval. This captures uncertainty over the item sample for \emph{these specific judges} on ChaosNLI; it does not generalize to other datasets or judge panels.

\subsection{Condorcet Null Model}
\label{sec:condorcet}

To translate $\neff$ into a concrete accuracy gap, we construct a Condorcet null model that simulates what majority-vote accuracy \emph{would be} if judges voted independently with the same error characteristics. Crucially, this tests \emph{conditional} independence: whether judges vote independently given the item's gold label and difficulty level. Some correlation from shared item difficulty is expected and accounted for; the gap measures dependence \emph{beyond} what difficulty explains.

\paragraph{Confusion-matrix calibration.}
For each judge $j$, we estimate the 3$\times$3 confusion matrix $P(\hat{y} = c' \mid y = c, j)$ from their labels on the 1{,}000 items. This captures class-specific error patterns (e.g., a judge that confuses entailment with neutral more than with contradiction).

\paragraph{Item-aware simulation.}
We stratify items into three difficulty bins by human entropy (terciles at the 33rd and 67th percentiles) and estimate per-judge, per-bin confusion matrices. We report results for 3 bins (terciles) as the default; \S4 reports sensitivity to 10 bins, and split-half cross-validation confirms minimal overfitting (Appendix~\ref{app:splithalf}). For each item, we run 10{,}000 Monte Carlo simulations: sample each judge's vote independently from the appropriate bin-specific confusion matrix given the item's gold label, compute majority vote, and record accuracy. This yields a predicted accuracy for each panel-entropy bin under the independence assumption, accounting for shared item difficulty.

\paragraph{Condorcet gap.}
The Condorcet gap is the difference between predicted (independent) and actual majority-vote accuracy, computed as a weighted average across panel-entropy bins (weighted by bin size). A negative gap indicates that actual accuracy falls \emph{below} the independent prediction. We compute a 95\% bootstrap confidence interval by resampling items 1{,}000 times and re-estimating the full pipeline for each resample.

\subsection{Statistical Tests}

\paragraph{Permutation omnibus test.}
To test whether the observed $\phiavg$ is significantly above chance, we conduct a stratified permutation test (10{,}000 permutations). Within each human-entropy stratum, we independently shuffle each judge's error vector, breaking inter-judge correlations while preserving per-judge error rates and the difficulty structure. We compute $\phiavg$ on each permuted dataset and report the fraction of permuted statistics that equal or exceed the observed value.

\paragraph{Per-bin binomial tests.}
For each discrete panel-entropy value, we conduct a one-sided binomial test of whether actual accuracy is significantly below the Condorcet prediction, with Wilson score confidence intervals. These per-bin tests are exploratory; the stratified permutation test serves as our primary omnibus significance test.

\section{Results}

\subsection{Effective Independence}

The 9-judge panel yields $\neff = 2.18$ with 95\% bootstrap CI $[2.07, 2.31]$ (Table~\ref{tab:headline}). The eigenvalue estimate ($\neff^{\text{eigen}} = 2.16$) closely matches, validating the Kish exchangeability assumption for this panel. The mean pairwise phi is $\phiavg = 0.391$ ($\sigma = 0.111$, range: $[0.161, 0.603]$), and the independence ratio is $\neff / k = 24.2\%$.

\begin{table}[t]
\centering
\small
\begin{tabular}{lc}
\toprule
\textbf{Metric} & \textbf{Value} \\
\midrule
Judges ($k$) & 9 \\
Families & 7 \\
Items ($n$) & 1{,}000 \\
\midrule
$\neff$ (Kish) & 2.18 [2.07, 2.31] \\
$\neff$ (eigenvalue) & 2.16 \\
$\lambda_{\max}$ & 4.17 \\
Mean $\phi$ & 0.391 $\pm$ 0.111 \\
Independence ratio & 24.2\% \\
\midrule
Panel accuracy & 72.0\% \\
Best individual (Qwen3-32B) & 71.8\% \\
Panel lift & $+$0.2pp \\
\midrule
Condorcet gap (weighted) & 22.0pp [19.5, 24.1] \\
Gap explained by difficulty & 6.8\% \\
Permutation $p$ & $< 10^{-4}$ \\
\bottomrule
\end{tabular}
\caption{Headline results. The 9-judge panel provides only 2.18 effective independent voters. The Condorcet gap measures the shortfall of actual accuracy below the Condorcet prediction for independent voters with the same per-judge error profiles. The panel's 0.2pp lift is within noise and tie-breaking margin (11 ties, 1.1\%).}
\label{tab:headline}
\end{table}

\begin{figure}[t]
    \centering
    \includegraphics[width=\columnwidth]{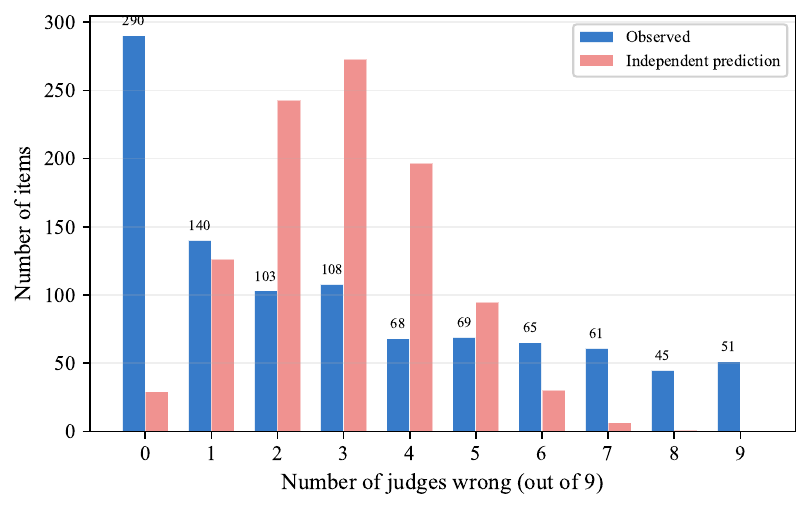}
    \caption{Distribution of errors per item. Under independence, errors would concentrate around 2--4 per item (right bars). The observed distribution (left bars) shows excess mass at the extremes --- 290 items with 0 errors and 51 with all 9 wrong (vs.\ $<$1 expected) --- the hallmark of correlated errors.}
    \label{fig:errors}
\end{figure}

The error distribution across items (Figure~\ref{fig:errors}) reveals the signature of correlated errors: 290 items (29\%) have all 9 judges correct and 51 (5.1\%) have all 9 wrong --- far more than any independence model predicts ($<$1). Over-prediction of \emph{contradiction} accounts for 51\% of all-wrong confusions despite comprising only 16.5\% of gold labels (Appendix~\ref{app:allwrong}).

\subsection{Condorcet Gap}

Majority-vote accuracy is 72.0\%, compared to the Condorcet prediction of approximately 94\% for the item-aware model. The weighted Condorcet gap is 22.0 percentage points (95\% CI: $[19.5, 24.1]$pp). Only 6.8\% of this gap is attributable to shared item difficulty; with 10 difficulty bins, the explained fraction rises to 13.5\% on MNLI but 66--87\% remains unexplained across all datasets. Split-half validation (ratio = 1.00) and the permutation test ($p < 10^{-4}$) confirm this is not overfitting.

The per-bin breakdown (Appendix~\ref{app:condorcet_fig}, Table~\ref{tab:calibration}) shows the gap is significant ($p < 0.05$) in 8 of 12 discrete panel-entropy levels. Even for unanimous items (panel entropy = 0, $n = 319$), accuracy is 90.9\% --- not the 99.99\% that Condorcet would predict for 9 independent voters each with $\sim$68\% accuracy.

\subsection{Permutation Test}

The permutation omnibus test yields $p < 10^{-4}$ (0 of 10{,}000 permutations reached the observed $\phiavg = 0.391$; permutation null: mean $= 0.060$, SD $= 0.005$, $z = 65.6$). This decisively rejects the null hypothesis that the observed inter-judge correlation is attributable to shared item difficulty alone.

\subsection{Scaling: $\neff(k)$ vs.\ $k$}

Figure~\ref{fig:scaling} shows how $\neff$ varies with panel size $k$ across all $\binom{9}{k}$ subsets (full data in Appendix~\ref{app:scaling_data}). The empirical curve closely tracks the Kish prediction $\neff(k) = k / (1 + (k-1) \cdot 0.391)$, with a hard asymptote at $1/\phiavg \approx 2.6$. The diminishing returns are severe: the first 5 judges contribute 90\% of the achievable independence ($\neff = 1.96$ vs.\ 2.18). Adding judges 6--9 provides only $+0.22$ effective votes.

\begin{figure}[t]
    \centering
    \includegraphics[width=\columnwidth]{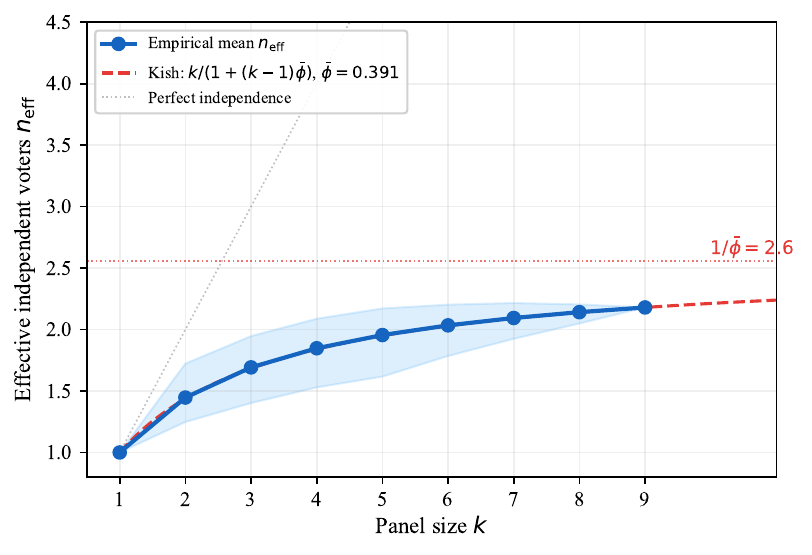}
    \caption{Effective independence $\neff$ as a function of panel size $k$. The empirical mean (blue circles) closely follows the Kish prediction (red dashes), far below the perfect-independence diagonal (gray). The shaded region shows the min--max range across all $\binom{9}{k}$ subsets. The asymptote at $1/\phiavg \approx 2.6$ means no panel of current models can exceed $\sim$2.6 effective independent votes.}
    \label{fig:scaling}
\end{figure}

\subsection{Cross-Dataset Replication}
\label{sec:snli}

To test whether the independence deficit generalizes beyond a single NLI source corpus, we replicate the full analysis on 1{,}000 ChaosNLI-SNLI items \citep{bowman2015large} (same 3-class task, different corpus) and 1{,}000 ChaosNLI-AlphaNLI items \citep{bhagavatula2020abductive} (2-class abductive reasoning, fundamentally different task type). Table~\ref{tab:crossdataset} summarizes the comparison.

\begin{table}[t]
\centering
\footnotesize
\setlength{\tabcolsep}{3pt}
\begin{tabular}{lccc}
\toprule
\textbf{Metric} & \textbf{MNLI} & \textbf{SNLI} & \textbf{AlphaNLI} \\
\midrule
Task type & 3-class NLI & 3-class NLI & 2-class abd. \\
$\neff$ (Kish) & 2.18\,[2.07, 2.31] & 2.35\,[2.21, 2.51] & 2.48\,[2.32, 2.69] \\
Mean $\phi$ & 0.391 & 0.354 & 0.328 \\
Panel acc. & 72.0\% & 77.7\% & 88.7\% \\
Best indiv. & 71.8\% & 84.2\% & 91.2\% \\
Panel lift & $+$0.2pp & $-$6.5pp & $-$2.5pp \\
Cond.\ gap (pp) & 22.0\,[19.5, 24.1] & 14.0\,[11.9, 16.1] & 7.6\,[6.0, 9.1] \\
Kripp.\ $\alpha$ & .550\,[.528, .573] & .546\,[.521, .568] & .577\,[.549, .601] \\
Human $\neff$ & 5.79 & 4.78 & 4.03 \\
Perm.\ $p$ & $< 10^{-4}$ & $< 10^{-4}$ & $< 10^{-4}$ \\
Split-half & 1.00 & 0.96 & 1.00 \\
\bottomrule
\end{tabular}
\caption{Cross-dataset comparison. All three datasets show $\neff \ll k$, significant Condorcet gaps, and negligible or negative panel lift. The independence deficit is remarkably stable ($\neff \approx 2.2$--$2.5$) despite varying task types, label sets, and base accuracy levels. Krippendorff's $\alpha < 0.667$ on all datasets, indicating only moderate inter-judge agreement by annotation science standards. Human $\neff$ is estimated under an exchangeability assumption by sampling from the aggregate ChaosNLI label distribution (see footnote in text).}
\label{tab:crossdataset}
\end{table}

The core finding replicates across all three datasets: $\neff$ remains in the narrow 2.2--2.5 range despite panel accuracy ranging from 72\% to 89\%, and the best individual judge matches or outperforms the panel in every case. On MNLI, the panel edges the best judge by a negligible 0.2pp; on SNLI and AlphaNLI, the best individual wins convincingly ($-$6.5pp and $-$2.5pp). The panel underperforms relative to the \emph{Condorcet prediction}, which already accounts for each judge's individual accuracy --- the gap is driven by correlated errors, not merely vote dilution. The AlphaNLI result is particularly noteworthy: a 2-class abductive reasoning task with fundamentally different cognitive demands, yet $\neff = 2.48$. Condorcet gaps decrease with base accuracy (22.0pp $\to$ 14.0pp $\to$ 7.6pp), as expected when higher accuracy leaves less room for correlated errors.

Krippendorff's $\alpha < 0.667$ on all datasets, indicating only moderate agreement by annotation standards \citep{krippendorff2011computing}. Human annotator panels achieve roughly $2\times$ higher $\neff$ ($4.0$--$5.8$ vs.\ LLMs' $2.2$--$2.5$), suggesting the deficit is specific to LLM judges.\footnote{Human $\neff$ is estimated by sampling 10 labels per item from the aggregate ChaosNLI distribution, treating annotators as exchangeable.}

Split-half cross-validation confirms no overfitting (ratios 0.96--1.00), and $\neff$ stabilizes by $N \approx 200$ (Appendix~\ref{app:convergence}).

\subsection{Robustness to Prompt, Temperature, and Task}
\label{sec:robustness}

To test whether the independence deficit is a prompt or decoding artifact, we re-run all 9 judges on the same 1{,}000 MNLI items under four variants: (1) \textbf{reframed} wording, (2) \textbf{reversed} label order, (3) \textbf{chain-of-thought} reasoning, and (4) temperature $T = 0.5$. We also evaluate on 1{,}000 RewardBench items \citep{lambert2024rewardbench} --- a pairwise preference task with deterministic gold labels, sampled via proportional stratified sampling across four categories, using the official MT-Bench pairwise judge prompt with A/B position randomization.

\begin{table}[t]
\centering
\small
\setlength{\tabcolsep}{3pt}
\begin{tabular}{lcccc}
\toprule
\textbf{Condition} & $\neff$ \textbf{[95\% CI]} & $\phiavg$ & \textbf{Panel} & \textbf{Gap} \\
\midrule
Baseline ($T{=}0$) & 2.18\,[2.07, 2.31] & .391 & 72.0\% & 22.0pp \\
Reframed prompt & 2.17\,[2.05, 2.30] & .394 & 72.5\% & 21.5pp \\
Reversed labels & 2.15\,[2.03, 2.27] & .399 & 72.9\% & 21.3pp \\
Chain-of-thought & 1.94\,[1.85, 2.04] & .456 & 69.2\% & 22.3pp \\
Temp $T{=}0.5$ & 2.17\,[2.06, 2.30] & .393 & 71.8\% & 21.9pp \\
\midrule
RewardBench & 1.99\,[1.83, 2.20] & .440 & 92.7\% & 6.8pp \\
\bottomrule
\end{tabular}
\caption{Robustness of $\neff$ across prompt variants and chain-of-thought (same 1{,}000 MNLI items, 9 judges) and a different task type (RewardBench: pairwise preference, 9 judges, 1{,}000 items). ``Gap'' is the Condorcet gap (predicted $-$ actual panel accuracy). $\neff$ is stable in the 1.94--2.18 range across all conditions; chain-of-thought increases correlation.}
\label{tab:robustness}
\end{table}

Table~\ref{tab:robustness} shows that $\neff$ is remarkably stable. Varying prompt wording, label ordering, and temperature has essentially no effect: $\neff$ ranges from 2.15 to 2.18, with overlapping 95\% bootstrap CIs. The reversed-label variant rules out position bias \citep{wang2024large} as a driver: the near-identical $\neff$ (2.15 vs.\ 2.18) confirms that correlation is robust to label ordering. Chain-of-thought actually \emph{increases} correlation ($\phiavg = .456$, $\neff = 1.94$) --- shared reasoning amplifies shared errors. This stability rules out prompt engineering artifacts.

On RewardBench --- a binary pairwise preference task with deterministic gold labels --- $\neff = 1.99$ [1.83, 2.20]. The smaller Condorcet gap (6.8pp vs.\ 21--22pp on MNLI) reflects higher panel accuracy (92.7\%), but $\phi = 0.44$ confirms high error correlation regardless of task type. Same-family correlation is larger on RewardBench ($+0.109$) than MNLI ($+0.047$). All 9 judges show residual A-preference despite the anti-bias prompt; the NLI results, structurally immune to position effects, confirm that correlation is not a position-bias artifact.

\section{Analysis and Discussion}

\subsection{Leave-One-Out: Which Judges Matter?}

Leave-one-out analysis (Appendix~\ref{app:supplementary}, Table~\ref{tab:loo}) reveals that herding is systemic: $\Delta\neff$ ranges narrowly from $-0.13$ to $+0.02$ across judges, with no single model driving the effect. Removing DeepSeek-V3 or Mistral Large 3 \emph{increases} $\neff$ (their errors are most correlated with the panel), while removing Llama 4 Scout decreases it the most.

Most strikingly, removing Gemini 2.5 Pro --- highly correlated with Claude ($\phi = 0.60$) and GPT-4o ($0.52$) --- \emph{increases} accuracy by 1.3pp (95\% CI $[+0.1, +2.6]$), and 6 of 9 removals improve accuracy. The three judges whose removal hurts (Maverick, Scout, Qwen3) include the two most individually accurate. That adding voters can \emph{hurt} is theoretically predicted under positive correlation \citep{austen1996information}, but has not previously been demonstrated in the LLM judge setting.

\subsection{Stratified and Subset Analyses}

When stratified by gold NLI class (Appendix~\ref{app:supplementary}, Table~\ref{tab:stratified}), herding is present across all three classes: $\neff$ ranges from 1.85 (contradiction, $\phiavg = 0.482$) to 2.40 (neutral). Even on the 179 ``easy'' items (17.9\%) where $\geq$80\% of human annotators agree, $\neff = 2.67$ --- higher than the full set but far from 9, ruling out the explanation that herding is merely a response to item ambiguity. Panel entropy correlates with human entropy ($\rho_s = 0.301$) and predicts majority-vote correctness ($r_{pb} = -0.342$). Among unanimous items ($n = 319$), accuracy is 90.9\%; with any disagreement ($n = 681$), it drops to 63.1\%. The 9.1\% error rate on unanimous items is dramatically higher than the $\sim$0.02\% that independence would predict.

\subsection{Same-Family vs.\ Cross-Family Correlation}
\label{sec:family}

Same-family pairs (OpenAI-OpenAI: $\phi = 0.437$; Meta-Meta: $\phi = 0.435$) are only slightly more correlated than the cross-family mean ($\phiavg_{\text{cross}} = 0.389$, difference = 0.047). The three highest-correlated pairs are all \emph{cross-family}: Claude $\times$ Gemini ($\phi = 0.603$), GPT-4o $\times$ Claude ($\phi = 0.588$), and Mistral $\times$ DeepSeek ($\phi = 0.564$). When restricted to one judge per family (7 judges, selecting the best in each), $\neff$ \emph{decreases} to 1.93 --- a selection effect where the best judges concentrate errors on the same hard items (full matrix in Appendix~\ref{app:phi}). Family diversity alone does not recover independence.

\subsection{Does Smarter Aggregation Help?}
\label{sec:aggregation}

A natural question is whether the Condorcet gap can be closed by replacing na\"ive majority voting with more sophisticated aggregation. We test three established methods: (1) \textbf{Dawid-Skene EM} \citep{dawid1979maximum}, which estimates per-judge confusion matrices and true label posteriors via expectation-maximization without access to gold labels; (2) \textbf{accuracy-weighted voting}, which weights each judge by their individual accuracy (using 5-fold cross-validation to avoid label leakage); and (3) \textbf{Markowitz-optimal weighting}, which selects weights to minimize correlated error via the inverse phi correlation matrix (also cross-validated). The latter two methods use gold labels for weight estimation --- giving them an \emph{oracle advantage} that would be unavailable in practice.

\begin{table}[t]
\centering
\footnotesize
\setlength{\tabcolsep}{2.5pt}
\begin{tabular}{lccccr}
\toprule
\textbf{Method} & \textbf{Orac.} & \textbf{MNLI} & \textbf{SNLI} & \textbf{Alpha.} & \textbf{RB} \\
\midrule
Majority vote & No & 72.0 & 77.7 & 88.7 & 92.7 \\
Dawid-Skene EM & No & 70.7 & 77.6 & 89.5 & 92.7 \\
Acc-weighted (CV) & Yes & 72.2 & 77.7 & 88.7 & 92.7 \\
\midrule
Best individual & --- & 71.8 & 84.2 & 91.2 & 95.5 \\
Condorcet pred. & --- & 94.0 & 91.7 & 96.3 & 99.5 \\
\bottomrule
\end{tabular}
\caption{Aggregation methods vs.\ the Condorcet gap. Even with oracle access to gold labels (accuracy-weighted, 5-fold CV), the best stable method closes at most 11\% of the gap across all four datasets. The best individual judge outperforms all aggregation methods on SNLI, AlphaNLI, and RewardBench (RB). Markowitz-optimal weighting is omitted from the main table due to instability (Appendix~\ref{app:aggregation}).}
\label{tab:aggregation}
\end{table}

Table~\ref{tab:aggregation} shows the results. On MNLI, accuracy-weighted voting (5-fold CV) achieves 72.2\% --- a gain of just 0.2pp over majority vote, closing less than 1\% of the 22.0pp Condorcet gap. Dawid-Skene actually \emph{underperforms} majority vote on MNLI (70.7\%), illustrating that unsupervised EM can misestimate error rates when judges are highly correlated. On AlphaNLI, Dawid-Skene closes 10.5\% of the gap --- the best stable result across all four datasets. On SNLI, AlphaNLI, and RewardBench, the best individual judge outperforms \emph{every} aggregation method, including those with oracle access. Note that identifying the best individual also requires oracle access to gold labels; the comparison highlights that even oracle-informed \emph{weighting} cannot overcome correlation. Markowitz-optimal (phi-optimal) weighting closes 20.6\% of the gap on RewardBench but underperforms majority vote on AlphaNLI, illustrating the instability of correlation-based weighting (Appendix~\ref{app:aggregation}).

With only $\sim$2.2 effective independent votes, no weighting scheme can extract a third independent perspective --- including calibrated soft voting \citep{ni2026reasoning, maiapolo2025bridging}. Our oracle-access stable methods close at most 11\% of the gap. Confounder-aware methods \citep{zhao2025care} face the same structural limit: same-family pairs are only marginally more correlated than cross-family ($+$0.047; \S\ref{sec:family}), and the three highest-correlated pairs are all cross-family.

\section{Conclusion}

We have applied the Kish effective sample size framework to LLM judge panels, providing the first measurement that directly connects inter-judge correlation to majority-vote reliability via Condorcet theory. The independence deficit ($\neff \approx 2.0$--$2.5$) is stable across three NLI datasets, three prompt variants, two temperature settings, and a pairwise preference task (RewardBench), confirming that the correlation is structural rather than an artifact of any particular experimental choice. Adding judges does not help: the panel matches or underperforms the best individual judge across all conditions. Established stable aggregation methods close at most 11\% of the Condorcet gap (unstable correlation-aware weighting reaches 21\% on one dataset but hurts on others), confirming that the bottleneck is in the inputs, not the algorithm.

These results have direct practical implications. Paying for 9 opinions but receiving the informational equivalent of $\sim$2 is a substantial inefficiency: a 5-judge panel already captures 90\% of achievable independence. Unanimous panel agreement carries far less weight than it appears --- our data show a 9.1\% error rate on unanimous items, vs.\ $\sim$0.02\% under independence. We recommend computing $\neff$ as a standard panel diagnostic: if $\neff / k < 0.5$, results should be treated with caution.

Our findings complement \citet{dorner2025limits} and \citet{jiang2025hivemind}. The path forward requires models that genuinely differ in how they process information --- not merely different brand names on similar architectures. The Kish formula makes progress measurable: halving $\phiavg$ from 0.39 to 0.20 would raise $\neff$ from 2.2 to 3.5, closing roughly half the Condorcet gap. Whether architecturally diverse models, specialist fine-tuning, or hybrid human-LLM panels can achieve this remains an open question.

\section*{Limitations}

\paragraph{Classification tasks.}
Our results are replicated across three ChaosNLI NLI datasets and a pairwise preference task (RewardBench), with consistent $\neff \approx 2.0$--$2.5$. The cross-task replication strengthens generalizability, but all four remain classification or binary preference tasks. The degree of inter-judge correlation may differ on open-ended generation evaluation or code review, where output structure differs fundamentally.

\paragraph{Gold standard validity.}
The 100-annotator majority label is our ground truth, but for high-entropy items, the majority label may represent a plurality preference rather than a ``correct'' answer. Appendix~\ref{app:distributional} reports distributional alignment metrics confirming the same pattern without reducing labels to binary accuracy.

\paragraph{Snapshot in time.}
Our results reflect a snapshot of current frontier models. Future models may exhibit lower correlation, but the \emph{framework} ($\neff$ and Condorcet gap) remains applicable.

\paragraph{Condorcet model calibration.}
The confusion matrices are estimated from the same items on which we measure the gap. Split-half cross-validation (Appendix~\ref{app:splithalf}) yields overfitting ratios of 0.96--1.00 across the three datasets, confirming negligible overfitting.

\paragraph{Prompt and decoding choices.}
We test four prompt variants (including chain-of-thought) and two temperature settings (\S\ref{sec:robustness}), finding $\neff$ stable in the 1.94--2.18 range. Chain-of-thought actually \emph{increases} correlation ($\neff = 1.94$). More radical prompt reformulations --- such as few-shot exemplars or persona-based prompting --- could in principle alter the correlation structure. We use RewardBench \citep{lambert2024rewardbench} rather than the more recent RewardBench~2 \citep{lambert2025rewardbench2} because the latter uses LLM-derived gold labels for several categories (e.g., GPT-4o and Claude consensus for Factuality), which would introduce circularity when evaluating LLM judges drawn from the same model families.

\paragraph{Bootstrap CI interpretation.}
Our bootstrap CI captures uncertainty over items for \emph{these specific judges}. It does not account for judge selection uncertainty --- a different panel might yield a different $\neff$.

\section*{Ethics Statement}
This work uses publicly available benchmark data (ChaosNLI) and commercial LLMs. No human subjects were recruited for this study. The ChaosNLI annotations were collected by \citet{nie2020chaosnli} and are publicly released. Our findings highlight limitations of LLM judge panels, which we believe serve the public interest by encouraging more careful deployment of automated evaluation systems. Claude (Anthropic) was used for writing assistance.

\section*{Data Availability}
ChaosNLI-MNLI is publicly available via HuggingFace (\texttt{metaeval/chaos-mnli-ambiguity}); ChaosNLI-SNLI and ChaosNLI-AlphaNLI are available from the ChaosNLI GitHub repository \citep{nie2020chaosnli}. RewardBench is available via HuggingFace (\texttt{allenai/reward-bench}; \citealt{lambert2024rewardbench}).

\bibliography{references}

\appendix

\section{NLI Classification Prompt}
\label{app:prompt}

All judges receive the following prompt at temperature 0.0:

\begin{quote}
\small
Given the following premise and hypothesis, determine the relationship between them.

Premise: \{premise\}

Hypothesis: \{hypothesis\}

What is the relationship? Reply with ONLY one word: ``entailment'', ``neutral'', or ``contradiction''.
\end{quote}

We use the canonical NLI label vocabulary (entailment, neutral, contradiction) rather than crowdsourcing-style phrasing (``definitely true'', ``might be true'', ``definitely not true'') because LLMs are trained primarily on NLI benchmark data that uses these terms; non-standard labels risk introducing an additional source of variance unrelated to the underlying judgment.

Parse failures are handled by deterministic hash-based random assignment to \{e, n, c\}, ensuring reproducibility and avoiding systematic bias.

\subsection{RewardBench Pairwise Preference Prompt}
\label{app:rb_prompt}

For RewardBench evaluation, we use the default pairwise judge prompt from the RewardBench codebase \citep{lambert2024rewardbench}, originally from MT-Bench \citep{zheng2023judging}. The RewardBench codebase provides several prompt variants for different judge architectures (e.g., Prometheus, OffsetBias, Atla); we use the default \texttt{MTBENCH\_V2} template, which is the standard for generative LLM-as-judge evaluation on the RewardBench leaderboard. The system message instructs judges to act as impartial evaluators, provide a short explanation, and output \texttt{[[A]]} or \texttt{[[B]]}. The user message presents the prompt and both responses with structured delimiters. All judges use temperature 0.0 and \texttt{max\_tokens}=4096.

\section{Phi Correlation Matrix}
\label{app:phi}

\begin{figure}[htbp]
    \centering
    \includegraphics[width=1.08\columnwidth]{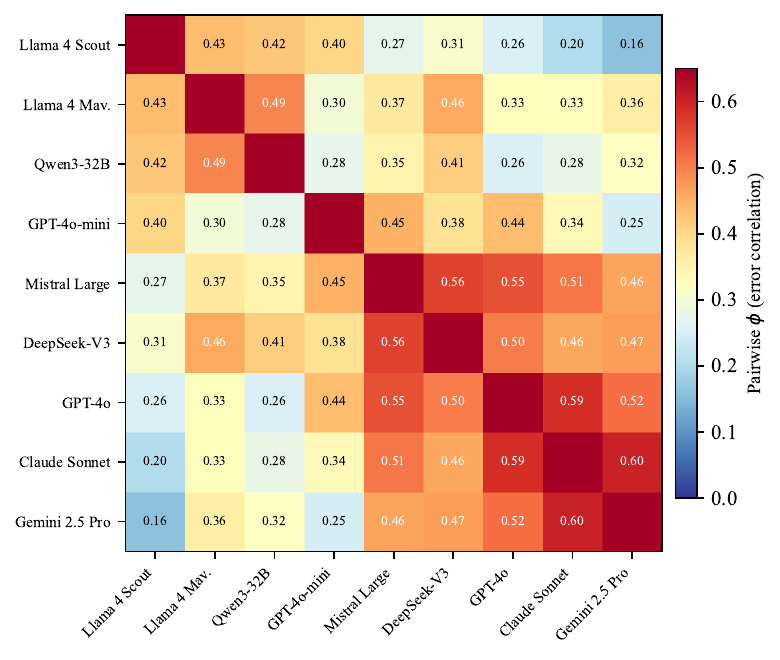}
    \caption{Pairwise phi correlation matrix (error correlation) for all 9 judges, ordered by hierarchical clustering. Cross-family pairs (e.g., Claude $\times$ Gemini, $\phi = 0.60$) can be as correlated as same-family pairs (OpenAI-OpenAI: $\phi = 0.437$; Meta-Meta: $\phi = 0.435$).}
    \label{fig:heatmap}
\end{figure}

Figure~\ref{fig:heatmap} shows the full $9 \times 9$ pairwise phi correlation matrix, ordered by hierarchical clustering.
Notable high-correlation pairs: Claude Sonnet $\times$ Gemini 2.5 Pro ($\phi = 0.603$), GPT-4o $\times$ Claude Sonnet ($\phi = 0.588$), Mistral Large 3 $\times$ DeepSeek-V3 ($\phi = 0.564$). Notable low-correlation pair: Gemini 2.5 Pro $\times$ Llama 4 Scout ($\phi = 0.161$).

\section{Condorcet Gap Visualization}
\label{app:condorcet_fig}

\begin{figure}[htbp]
    \centering
    \includegraphics[width=\columnwidth]{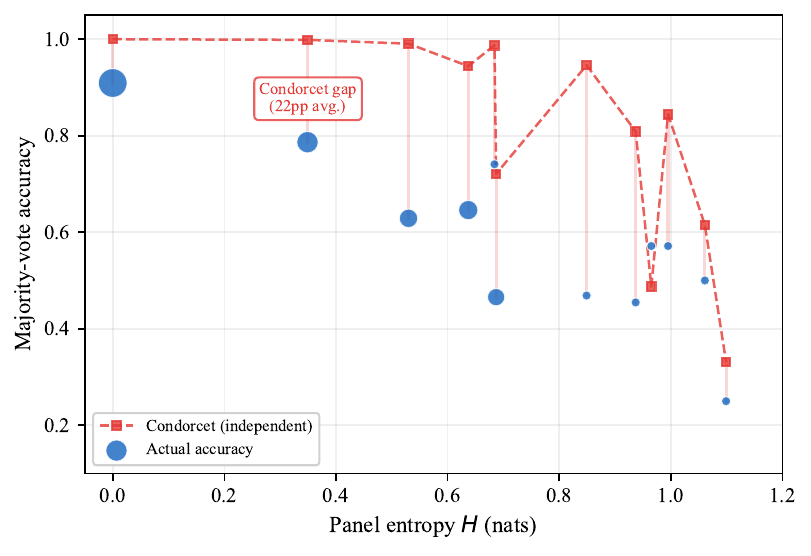}
    \caption{Condorcet gap by panel entropy. Blue circles: actual majority-vote accuracy (size proportional to bin $n$). Red squares: Condorcet predicted accuracy under the independence assumption. The shaded vertical lines highlight the gap. Average weighted gap: 22.0pp.}
    \label{fig:condorcet}
\end{figure}

Figure~\ref{fig:condorcet} shows the per-bin Condorcet gap. The vertical distance between the Condorcet prediction (red) and actual accuracy (blue) visualizes the independence deficit: the gap is largest for moderate-entropy items where independent voting should still yield high accuracy but actual voting fails.

\begin{table}[htbp]

\centering
\small
\begin{tabular}{rrrrrr}
\toprule
\textbf{H\textsubscript{panel}} & \textbf{$n$} & \textbf{Actual} & \textbf{Cond.} & \textbf{Gap} & \textbf{$p$} \\
\midrule
0.000 & 319 & .909 & 1.000 & $-$.091 & $<$.001 \\
0.349 & 178 & .787 & .999 & $-$.212 & $<$.001 \\
0.530 & 132 & .629 & .991 & $-$.362 & $<$.001 \\
0.637 & 144 & .646 & .944 & $-$.298 & $<$.001 \\
0.684 & 27 & .741 & .987 & $-$.246 & $<$.001 \\
0.687 & 116 & .466 & .721 & $-$.255 & $<$.001 \\
0.849 & 32 & .469 & .946 & $-$.477 & $<$.001 \\
0.937 & 22 & .455 & .809 & $-$.354 & $<$.001 \\
0.965 & 7 & .571 & .488 & $+$.084 & .793 \\
0.995 & 7 & .571 & .844 & $-$.273 & .081 \\
1.061 & 12 & .500 & .615 & $-$.115 & .297 \\
\bottomrule
\end{tabular}
\caption{Per-bin Condorcet gap analysis. H\textsubscript{panel} = discrete panel entropy (nats). Cond.\ = item-aware Condorcet predicted accuracy. Gap = actual $-$ predicted. $p$-values from one-sided binomial tests. One additional bin with $n < 5$ is omitted.}
\label{tab:calibration}
\end{table}

\section{Distributional Alignment Analysis}
\label{app:distributional}

Our main analyses use binary accuracy (match to the 100-annotator majority label). Since ChaosNLI involves 3-way classification with genuine human disagreement, we verify our findings using distributional metrics that do not reduce labels to right/wrong.

For each item, we compare the panel's label distribution (9 votes over \{e, n, c\}) to the human distribution (100 annotations) using total variation (TV) distance and symmetric KL divergence. Table~\ref{tab:distributional} reports results stratified by human-entropy tercile.

\begin{table}[htbp]

\centering
\small
\begin{tabular}{lccc}
\toprule
\textbf{Tercile} & \textbf{$n$} & \textbf{Mean TV} & \textbf{Mean sym-KL} \\
\midrule
Low (easy) & 334 & 0.192 $\pm$ 0.142 & 1.25 \\
Medium & 333 & 0.261 $\pm$ 0.151 & 1.64 \\
High (hard) & 333 & 0.345 $\pm$ 0.157 & 2.42 \\
\midrule
Overall & 1{,}000 & 0.266 $\pm$ 0.163 & 1.77 \\
\bottomrule
\end{tabular}
\caption{Distributional alignment between panel and human label distributions. TV distance ranges from 0 (identical) to 1 (disjoint). Higher human entropy (harder items) yields larger distributional divergence, confirming the pattern observed with binary accuracy.}
\label{tab:distributional}
\end{table}

TV distance correlates strongly with human entropy ($\rho_s = 0.434$, $p < 10^{-46}$): the panel's distributional misalignment grows with item ambiguity. This parallels the Condorcet gap finding (\S4.2) but does not depend on reducing the 3-class problem to binary accuracy. On low-entropy items where accuracy is a clean metric (mean TV $= 0.192$), the panel is reasonably well-calibrated; on high-entropy items (mean TV $= 0.345$), the panel's distribution diverges substantially from the human distribution --- consistent with correlated judges collapsing onto a single class rather than reflecting the human spread.

\section{Split-Half Condorcet Validation}
\label{app:splithalf}

To verify that our Condorcet gap estimate is not inflated by overfitting, we perform split-half cross-validation. We split the 1{,}000 items into two halves (500 each), stratified by human-entropy tercile to preserve difficulty balance. We estimate per-judge, per-bin confusion matrices on half~A, simulate the Condorcet prediction on half~B, and vice versa.

The cross-validated weighted Condorcet gap is 21.9pp for MNLI, closely matching the in-sample estimate of 22.0pp (overfitting ratio = 0.997). Across all three datasets, the overfitting ratios are 0.997 (MNLI), 0.960 (SNLI), and 1.000 (AlphaNLI) --- all near unity, confirming that with 1{,}000 items and 9-parameter confusion matrices per judge, the gap estimate is stable and not an artifact of in-sample fitting.

\section{Sample Size Convergence}
\label{app:convergence}

\begin{figure}[htbp]
    \centering
    \includegraphics[width=\columnwidth]{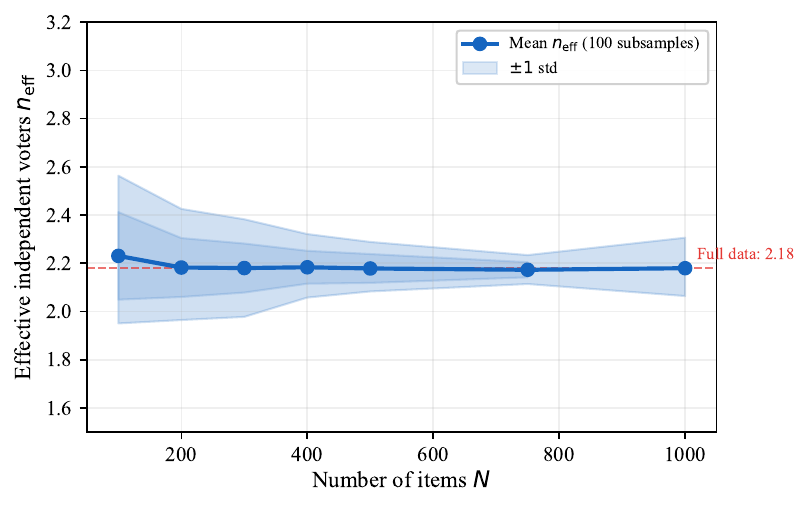}
    \caption{$\neff$ as a function of sample size $N$. For $N < 1{,}000$: mean over 100 independent stratified subsamples, with 95\% interval (shaded) and $\pm 1$ std (darker band). For $N = 1{,}000$: bootstrap CI. The estimate stabilizes by $N \approx 200$--300, confirming that 1{,}000 items provide a reliable measurement.}
    \label{fig:convergence}
\end{figure}

A natural concern is whether 1{,}000 items provide a stable estimate of $\neff$. Figure~\ref{fig:convergence} shows $\neff$ computed on entropy-stratified subsamples of increasing size ($N \in \{100, 200, 300, 400, 500, 750, 1000\}$). For each $N < 1{,}000$, we draw 100 independent stratified subsamples and report the mean $\neff$ with 2.5th--97.5th percentile bands; for $N = 1{,}000$ (the full dataset), we report the bootstrap CI (10{,}000 resamples). Using multiple independent subsamples per $N$ avoids the non-monotonic artifacts that a single random draw can produce. The estimate stabilizes by $N \approx 200$--300, and the interval narrows substantially by $N = 500$. This confirms that our 1{,}000-item sample provides a reliable measurement.

\section{Scaling Curve Data}
\label{app:scaling_data}

\begin{table}[htbp]
\centering
\small
\begin{tabular}{cccc}
\toprule
\textbf{$k$} & \textbf{Mean $\neff$} & \textbf{Kish pred.} & \textbf{Max $\neff$} \\
\midrule
2 & 1.45 & 1.44 & 1.72 \\
3 & 1.69 & 1.68 & 1.95 \\
4 & 1.85 & 1.84 & 2.09 \\
5 & 1.96 & 1.95 & 2.17 \\
6 & 2.03 & 2.03 & 2.20 \\
7 & 2.09 & 2.09 & 2.22 \\
8 & 2.14 & 2.14 & 2.20 \\
9 & 2.18 & 2.18 & --- \\
\midrule
$\infty$ & --- & 2.56 & --- \\
\bottomrule
\end{tabular}
\caption{$\neff(k)$ scaling curve. Mean $\neff$ across all $\binom{9}{k}$ subsets closely matches the Kish formula with $\phiavg = 0.391$. Asymptote at $1/\phiavg \approx 2.56$.}
\label{tab:scaling}
\end{table}

\section{Supplementary Analysis Tables}
\label{app:supplementary}

Tables~\ref{tab:loo} and~\ref{tab:stratified} provide the full leave-one-out and per-class stratification data referenced in \S5.

\begin{table}[htbp]
\centering
\small
\begin{tabular}{lrrrr}
\toprule
\textbf{Judge} & \textbf{Fam.} & \textbf{$\Delta\neff$} & \textbf{Acc w/o} & \textbf{$\Delta$Acc} \\
\midrule
DeepSeek-V3 & DS & $+$0.02 & .725 & $+$.005 \\
Mistral Large 3 & Mis & $+$0.02 & .728 & $+$.008 \\
GPT-4o & OAI & $+$0.01 & .726 & $+$.006 \\
Claude Sonnet & Ant & $-$0.01 & .729 & $+$.009 \\
Gemini 2.5 Pro & Goo & $-$0.04 & .733 & $+$.013 \\
Llama 4 Mav. & Meta & $-$0.05 & .717 & $-$.003 \\
GPT-4o-mini & OAI & $-$0.08 & .727 & $+$.007 \\
Qwen3-32B & Ali & $-$0.08 & .715 & $-$.005 \\
Llama 4 Scout & Meta & $-$0.13 & .719 & $-$.001 \\
\bottomrule
\end{tabular}
\caption{Leave-one-out analysis. $\Delta\neff$ is the change in $\neff$ when a judge is removed. Sorted by $\Delta\neff$. Removing 6 of 9 judges improves accuracy; only 3 removals hurt (Maverick, Qwen3, Scout --- including the two most individually accurate judges).}
\label{tab:loo}
\end{table}

\begin{table}[htbp]
\centering
\small
\begin{tabular}{lccc}
\toprule
\textbf{Class} & \textbf{$n$} & $\neff$ & $\phiavg$ \\
\midrule
Entailment & 476 & 1.90 & 0.466 \\
Neutral & 359 & 2.40 & 0.343 \\
Contradiction & 165 & 1.85 & 0.482 \\
\bottomrule
\end{tabular}
\caption{$\neff$ stratified by gold NLI class. Herding is present across all classes but strongest for contradiction ($\phiavg = 0.482$) and entailment ($\phiavg = 0.466$).}
\label{tab:stratified}
\end{table}

\section{Aggregation Method Details}
\label{app:aggregation}

Table~\ref{tab:aggregation_full} shows the full aggregation comparison including phi-optimal (Markowitz) weighting, which is omitted from the main table due to instability. Phi-optimal weighting selects judge weights via the inverse phi correlation matrix to minimize correlated error (analogous to Markowitz portfolio optimization). While it achieves the highest accuracy on MNLI (72.4\%) and RewardBench (94.1\%, closing 20.6\% of the gap), it \emph{hurts} performance on AlphaNLI (86.2\%, below MV), suggesting overfitting to the correlation structure. Cross-validated estimates reduce this instability but confirm the same conclusion: no method meaningfully closes the Condorcet gap.

\begin{table}[t]
\nolinenumbers\centering
\small
\setlength{\tabcolsep}{3pt}
\begin{tabular}{lcccccc}
\toprule
\textbf{Method} & \textbf{Orac.} & \textbf{CV} & \textbf{MNLI} & \textbf{SNLI} & \textbf{AlphaNLI} & \textbf{RB} \\
\midrule
Majority vote & No & --- & 72.0 & 77.7 & 88.7 & 92.7 \\
Dawid-Skene & No & --- & 70.7 & 77.6 & 89.5 & 92.7 \\
Acc-weighted & Yes & Yes & 72.2 & 77.7 & 88.7 & 92.7 \\
Phi-optimal & Yes & Yes & 72.4 & 78.4 & 86.2 & 94.1 \\
\midrule
Best indiv. & --- & --- & 71.8 & 84.2 & 91.2 & 95.5 \\
Condorcet & --- & --- & 94.0 & 91.7 & 96.3 & 99.5 \\
\bottomrule
\end{tabular}
\caption{Full aggregation comparison including phi-optimal (Markowitz) weighting across all four datasets (5-fold CV). Phi-optimal achieves the highest MNLI accuracy but hurts AlphaNLI (86.2\%, below MV), illustrating instability with correlated judges. On RewardBench (RB), it closes 20.6\% of the gap --- the highest across datasets --- but the best individual judge (Claude Sonnet 4.5, 95.5\%) still dominates.}
\label{tab:aggregation_full}
\end{table}

\section{All-Wrong Item Analysis}
\label{app:allwrong}

Table~\ref{tab:allwrong} breaks down the 51 items where all 9 judges are wrong.

\begin{table}[t]
\centering
\nolinenumbers\small
\begin{tabular}{lrr}
\toprule
\textbf{Category} & \textbf{Count} & \textbf{Frac.} \\
\midrule
\multicolumn{3}{l}{\emph{By human-entropy tercile}} \\
\quad Low (H $< 0.33$p) & 5 & 9.8\% \\
\quad Medium & 10 & 19.6\% \\
\quad High (H $> 0.67$p) & 36 & 70.6\% \\
\midrule
\multicolumn{3}{l}{\emph{By panel error type}} \\
\quad Biased ($\geq$50\% human maj.) & 29 & 56.9\% \\
\quad Ambiguous ($<$50\% majority) & 22 & 43.1\% \\
\midrule
\multicolumn{3}{l}{\emph{By confusion direction}} \\
\quad n $\to$ c & 14 & 27.5\% \\
\quad e $\to$ c & 12 & 23.5\% \\
\quad e $\to$ n & 11 & 21.6\% \\
\quad c $\to$ n & 7 & 13.7\% \\
\quad Other & 7 & 13.7\% \\
\bottomrule
\end{tabular}
\caption{Breakdown of the 51 all-wrong items (MNLI). Over-prediction of \emph{contradiction} accounts for 51\% of failures (n$\to$c + e$\to$c). Items where all 9 judges choose the same wrong label despite moderate human agreement highlight shared systematic biases. Mean human support for the panel's wrong answer across all 51 items is 35.3\%.}
\label{tab:allwrong}
\end{table}

\section{Tie-Breaking}
\label{app:tiebreaks}

\paragraph{Gold-label ties (100 annotators).}
With 100 annotators, ties are possible: 28 items across our three 1{,}000-item datasets have exactly tied top labels (e.g., 49n/49c/2e). The ChaosNLI dataset resolves these via Python \texttt{Counter} insertion order --- an undocumented artifact, not an intentional design choice. We verified that our results are robust: flipping all 28 tied items to the alternative label changes panel accuracy by at most $\pm$0.7pp and $\neff$ by at most 1.3\%, both well within bootstrap confidence intervals.

\paragraph{Majority-vote ties (9 judges).}
With 9 judges (an odd number), ties are impossible on 2-class tasks (AlphaNLI, RewardBench). On 3-class tasks (MNLI, SNLI), ties can still occur in 3-3-3 or 4-4-1 configurations, but are rare (MNLI: 11 ties or 1.1\%; SNLI: 4 ties or 0.4\%). These are broken via deterministic SHA-256 hashing of the item index and vote sequence, ensuring reproducibility and avoiding insertion-order bias. Monte Carlo simulations use random tie-breaking (seeded RNG) to correctly average over tie outcomes.
Tie frequencies: MNLI 11/1{,}000 (1.1\%; 4 at 3-3-3, 7 at 4-4-1), SNLI 4/1{,}000 (0.4\%; all 4-4-1), AlphaNLI 0/1{,}000 (0.0\%; ties impossible with 9 judges on a 2-class task), RewardBench 0/1{,}000 (0.0\%).

\end{document}